\documentclass{article}

\usepackage[preprint]{neurips_2025}
\usepackage{algorithm}
\usepackage{algpseudocode}
\usepackage{amssymb}
\usepackage{soul}

\usepackage[utf8]{inputenc} 
\usepackage[T1]{fontenc}    
\usepackage{hyperref}       
\usepackage{url}            
\usepackage{booktabs}       
\usepackage{amsfonts}       
\usepackage{nicefrac}       
\usepackage{microtype}      
\usepackage{xcolor}         
\usepackage{graphicx}
\usepackage{subcaption}
\usepackage{xspace}

\newcommand{\methodname}{\textsc{VideoWeave}\xspace}

\title{\methodname: A Data-Centric Approach \\ for Efficient Video Understanding}

\author{%
Zane Durante$^{1}$ 
\qquad
Silky Singh$^{1}$
\qquad
Arpandeep Khatua$^{1}$
\qquad
Shobhit Agarwal$^{1}$
\\
\textbf{Reuben Tan}$^{2}$
\qquad
\textbf{Yong Jae Lee}$^{3}$
\qquad
\textbf{Jianfeng Gao}$^{2}$
\qquad 
\textbf{Ehsan Adeli}$^{1}$
\qquad
\textbf{Li Fei-Fei}$^{1}$
\\
\\
$^{1}$ Stanford University \quad $^{2}$ Microsoft Research \quad $^{3}$ University of Wisconsin - Madison
}

\begin{document}

\maketitle

\begin{abstract}

Training video–language models is often prohibitively expensive due to the high cost of processing long frame sequences and the limited availability of annotated long videos. We present \methodname, a simple yet effective approach to improve data efficiency by constructing \textit{synthetic long-context} training samples that splice together short, captioned videos from existing datasets. Rather than modifying model architectures or optimization objectives, \methodname reorganizes available video–text pairs to expand temporal diversity within fixed compute. We systematically study how different data composition strategies like random versus visually clustered splicing and caption enrichment affect downstream performance on downstream video question answering. Under identical compute constraints, models trained with \methodname achieve higher accuracy than conventional video finetuning. Our results highlight that reorganizing training data, rather than altering architectures, may offer a simple and scalable path for training video–language models. We link our code for all experiments  \href{https://github.com/sagarwal02/videoweave}{here}.

\end{abstract}
\section{Introduction}

Vision-language models (VLMs)~\cite{radford2021learning, li_blip_2022, alayrac2022flamingo, yang2023dawn, zhang2023gpt4v, openai2023gpt4v} have demonstrated significant success in image understanding tasks, including captioning and visual question answering \cite{zhang2024vision, ozdemir2024enhancing, zhou2020unified, li2025survey, zhou2023vision+}. These advancements rely on large-scale image-text datasets that are used to train advanced multimodal encoders that learn rich representations of visual content. Although researchers have extended VLMs to video data with some success, the video domain presents unique challenges. One critical bottleneck is the limited availability of video-text training data compared to image-text pairs \cite{tan2024vidgen, wang2023internvid, yu2023celebv}. Current video-text datasets are orders of magnitude smaller than their image counterparts in both scale and diversity \cite{Bain21, schuhmann2022laion}. This scarcity is largely due to the prohibitive cost of annotating long videos, as annotators must watch extended footage to provide detailed captions or question-answer pairs. For example, several video datasets containing relatively short video clips (<1 minute in length) have been released that contain over 10 million training samples~\cite{Bain21,wanginternvid}. In contrast, recent benchmarks like HourVideo~\cite{chandrasegaran2024hourvideo}, VideoMME~\cite{fu2024video}, etc. feature videos up to 1 hour in length, and have highlighted performance gaps in state-of-the-art models, particularly for temporal reasoning tasks \cite{wu2024longvideobench, mavideo, ranasinghe2024understanding, li2025benchmark}. While these benchmarks effectively measure progress in models' long-form visual processing capabilities, they do not provide sufficient training data for models to develop these capabilities in the first place.

In addition to a lack of long-context training data, video-language models suffer from significantly higher compute requirements due to the need to sample multiple video frames per input rather than only a single image.  To address both challenges, we revisit how existing video–text data is organized and propose a data-centric framework that expands temporal diversity while maintaining fixed compute and annotation budgets.

In this work, we present \methodname, a simple yet effective approach to improve data efficiency by constructing \textit{synthetic long-context} training samples that splice together short, captioned clips. Our method concatenates existing short video segments into continuous streams, creating synthetic long videos without requiring additional human annotation. Specifically, our method addresses the data scarcity problem in long video understanding without requiring costly new annotations. Through our approach, we aim to enable more effective training of video language models for tasks requiring extended temporal comprehension. Our approach generates training data with more temporal and visual complexity that (1) allows for leveraging a larger  training set with fewer training iterations, and (2) outperforms standard single-video finetuning on the challenging VideoMME benchmark.

\section{Methodology}

\begin{figure*}[t]
    \centering
    \includegraphics[width=0.8\textwidth]{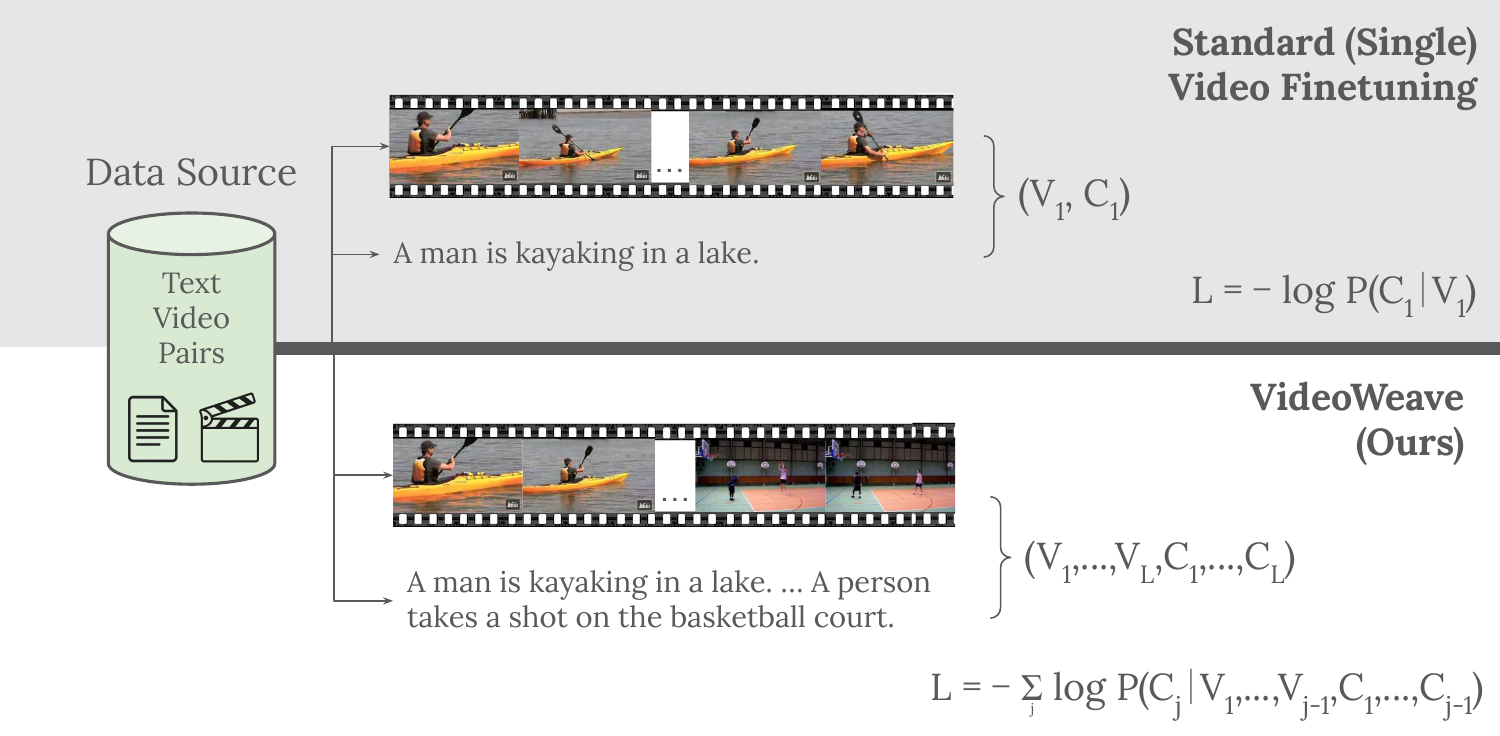}
    \caption{Our method \methodname combines several videos together to form a single input sample.  Our method is architecture and task agnostic, and is parametrized by $L$, the number of videos to use for each input pair. In this figure, we visualize $L=2$, representing two distinct video/caption pairs.}
    \label{fig:main_figure}
\end{figure*}


Motivated by the need for compute-efficient methods for training on long-context training data, we synthesize extended temporal contexts by concatenating multiple short, captioned video segments drawn from existing datasets. This approach allows us to effectively downsize the amount of training data by simultaneously decreasing the number of frames sampled from each video, thereby significantly decreasing the number of training steps required for convergence. In the following sections, we describe in detail our data curation pipeline and the overall architecture setup.





\subsection{Video Training Data Curation}
For all experiments, we use the WebVid-10M dataset \cite{Bain21} since it is a large scale dataset containing 10M video-caption pairs. Due to its large size, we can extensively evaluate our method across various training dataset sizes. We create multiple subsets of videos (smaller datasets of size 10K, 20K, ..., 160K) from WebVid-10M such that larger datasets are strict supersets of the smaller datasets, appropriately simulating the process of collecting more data from an identical data source. For our base method, we use the original, unaltered captions from WebVid along with sampling a fixed number of frames from each video.  To construct a single input sample, we construct a sequence of visual frames $V_t$ and a sequence of captions $T_n$, where $t=1,2,..T$ represents the indices for the frame set and $n=1,2,..N$ represents the indices of the video and caption pairs.  Our constructed input sample consists of $(V_1, V_2, ..., V_T,T_1, T_2, ..., T_N)$.  The specific values of $t$ and $n$ depend on the number of video-caption pairs ($N$) and the number of frames sampled from each video. We note that during training and evaluation, we assume a fixed number of frames as input to our model following previous works \cite{bertasius2021space,maaz2024video,wang2024tarsier}. For the purposes of our experiments, we set $T = 16$. To best simulate compute-bound training scenarios, we only train for a single epoch such that each input sample contains novel videos and text targets.

\subsection{Video-Language Model Architecture}

In this section, we detail how we modify existing image-based VLMs to take in video as input. Our approach for video splicing requires minimal changes to existing encoder-decoder vision-language model architectures. Typically there are three main components in a VLM: 1) the LLM backbone, 2) the vision encoder, and 3) a connector/bridge module (commonly a linear projection or MLP). However, by default the vision encoder is limited to processing a single image. In order to effectively leverage pretrained image-based VLMs, we simply encode each video frame ($V_1, V_2, ..., V_T$) independently. To process all frames at once, we squeeze the temporal dimension into the batch dimension (say, B) resulting in an input of shape (B * T, C, H, W). The input is then passed to the VLM as would a batch of image, and the output is reshaped to produce embeddings of shape (B, T, ...). 

The advantage of this approach is that we can still use image-based vision encoders, and can easily finetune further after initializing from existing checkpoints.  This approach is also commonly used in practice~\cite{wang2024tarsier} and introduces minimal architectural changes to existing VLMs.  For our implementation, we modify Prismatic-VLMs \cite{karamcheti2024prismatic} to handle video inputs and use a CLIP-ViT-L/14 \cite{radford2021learning} vision encoder with a LLaMA-2 \cite{touvron2023llama} decoder and a GELU MLP connector.  For all other hyperparameters and training recipe details, we follow Karamcheti et al. \citep{karamcheti2024prismatic}.

In our experimental setting, the inputs are a set of video frames and a text prompt. Unless specified otherwise, we use the following text prompt: ``Describe what is happening in the video.'' for all caption-based experiments. Additionally, we sample frames uniformly at random during training and test time. A practical concern arises when we use LLM backbones with smaller context lengths, the number of visual tokens are too large to fit within the context length of the LLM. To overcome this issue and maintain a fair comparison with image-based VLMs, we use RoPE~\cite{liu2024scalinglawsropebasedextrapolation} scaling to increase the context lengths. To train on 16 video frames (our default setting), we used a rope scaling factor of 3.0.






\subsection{Random Selection}
To set an initial baseline to understand how well video splicing can facilitate VLM video finetuning, we randomly select videos to group for each input sample.  We denote the number of videos to use per input as $L$.  We test $L=\{1,2,4,8,16\}$ ($L=1$ corresponds to standard single-video finetuning).  When splicing, we use the original captions $C_n$ and simply append an empty space `` " between captions. Despite it's simplicity, we found that random selection is an extremely strong baseline for video splicing, outperforming single-video finetuning and more complex strategies explored in Section \ref{sec:clustering} and Section \ref{sec:captioning}.

\subsection{Video Clustering}
\label{sec:clustering}


One possible hypothesis for constructing more meaningful input samples using video splicing is to choose video data that is visually similar.   Given a dataset $\mathcal{D}$ consisting of video-caption pairs; videos $V_n$ and corresponding captions $C_n$, we explore clustering and grouping videos together based on their visual feature similarity. 
By first clustering our input videos, we can ensure that for each input sample, we extract frames from a single cluster during training which will have more similar scenes. We note that the number of videos used in each input sample is a hyperparameter and that many standard clustering algorithms cannot enforce clusters of a fixed-size. In order to alleviate this issue, we modified the K-means algorithm~\cite{Jin2010} to achieve this objective as shown in Algorithm \ref{alg:clustering}. 

\begin{figure*}[h]
    \centering
    \includegraphics[width=1.0\textwidth]{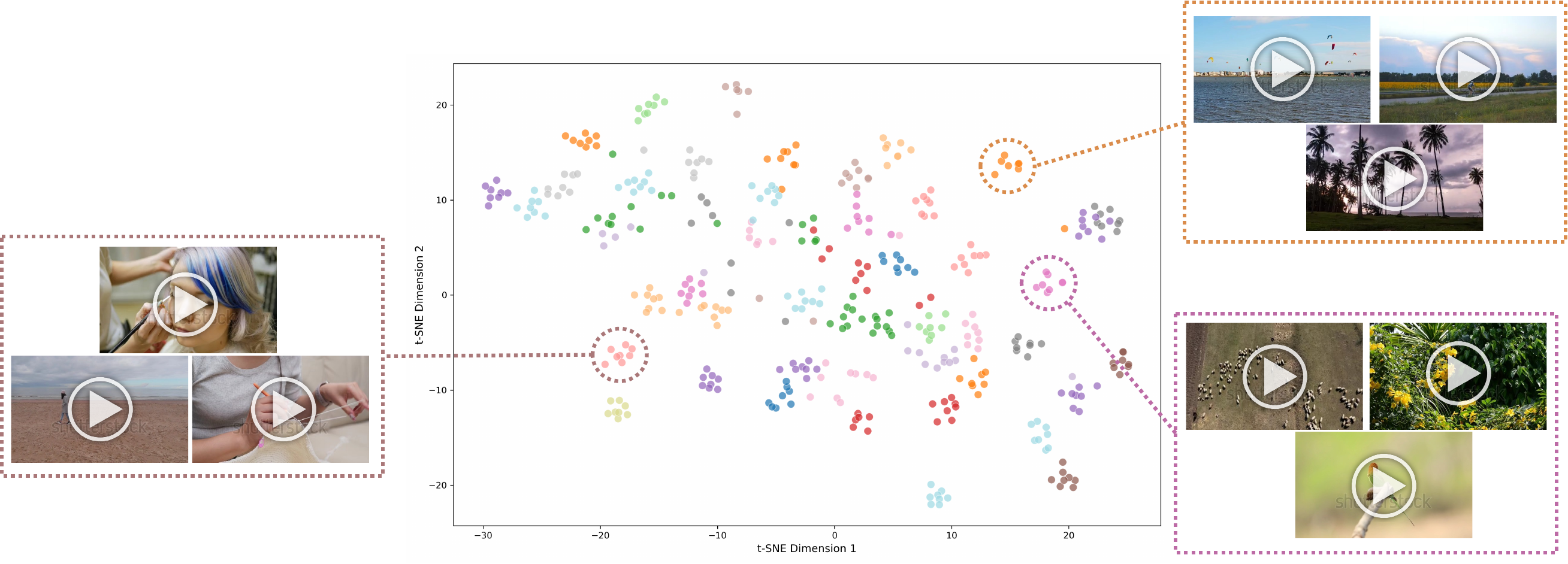}
    \caption{We visualize clusters generated by our modified K-means algorithm.  For the experiments described in Section \ref{sec:clustering}, we construct input samples by using video clips within a single cluster.}
    \label{fig:cluster-visualization}
\end{figure*}

To featurize each video, we take the average of CLIP-ViT-L/14 embeddings of uniformly sampled frames (say, $M$) from a video. These embeddings are then clustered using our modified K-means algorithm presented in Algorithm~\ref{alg:clustering}. In this work, we use CLIP~\cite{radford2021learning} as our image encoder $\mathcal{E}$ and $M$ is set to 16 by default. However, in practice, clustering similar videos together also has a downside -- the model sees a set of similar frames and could default to captioning the entire sequence based on first few frames. It doesn't force the model to understand what's happening in each frame. To tackle this, we propose uniform random sampling strategy. We put together randomly sampled videos in a cluster. This strategy provides enough variance to each data point for a coherent understanding of the video content.  We visualize clusters from a subset of WebVid10M in Figure \ref{fig:cluster-visualization}.

\begin{algorithm}
\caption{Modified K-means clustering algorithm}
\label{alg:clustering}
\begin{algorithmic}[1]
\Require Dataset $X$ with $n$ points, number of clusters $K$, $d$ data points per cluster
\Ensure Cluster assignments and final centroids
\State Initialize $K$ centroids $\mu_1, \mu_2, \dots, \mu_K$
\Repeat
    \For{each point $x_i \in X$}
        \State Assign $x_i$ to the nearest cluster $k$ such that 
        $\sum_{j=1}^{n} \mathbf{1}[c_j = k] < d$:
        \[
        c_i \gets \arg\min_{k} \|x_i - \mu_k\|^2
        \]
    \EndFor
    \For{each cluster $k = 1,\dots,K$}
        \State Update $\mu_k \gets \frac{1}{|C_k|}\sum_{x_i \in C_k} x_i$
    \EndFor
\Until{cluster assignments do not change}
\end{algorithmic}
\end{algorithm}

\subsection{Caption Generation}
\label{sec:captioning}





\begin{figure*}[h]
    \centering
    \includegraphics[width=1.0\textwidth]{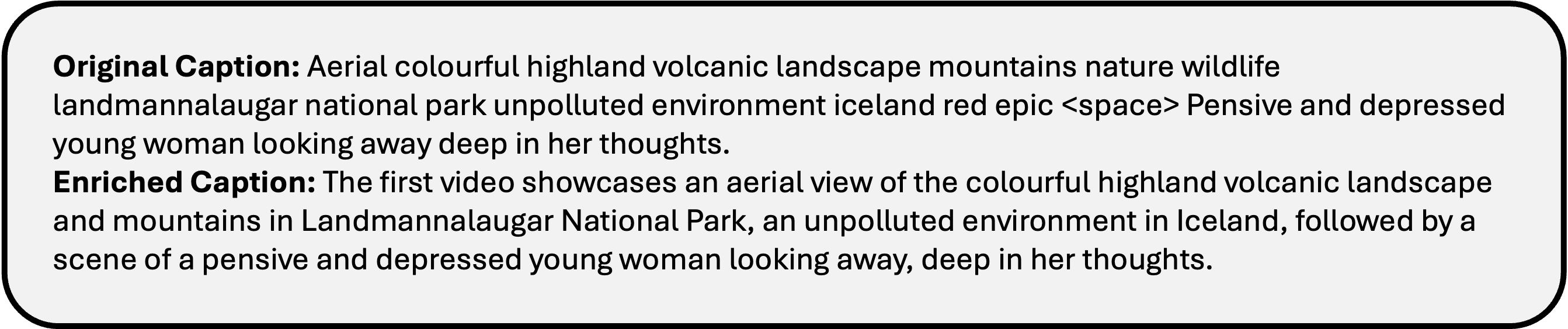}
    \caption{We show an example caption-caption generation pair where GPT-4o-mini modifies our naive caption into a unified, cohesive video caption.}
    \label{fig:caption}
\end{figure*}

We also explored leveraging language models to generate captions for the spliced videos. The target output for the language model is to produce a single caption describing all the videos in the input. It is common to prompt an LLM like GPT-4~\cite{achiam2023gpt} to generate question-answer pairs or summarizations without any visual baseline \cite{li2024llava} when using synthetic data to finetune VLMs. We leverage a similar approach -- for clustered videos with captions joined by a whitespace, we prompt GPT-4o-mini to rewrite the input captions into more cohesive captions that retain the original captions' semantic information.  Overall, we found that our enriched captions performed significantly worse than simply using the original WebVid-10M video captions. As a result, we use the joined captions in our experiments. Figure~\ref{fig:caption} shows an example output of an enriched caption.

\section{Experiments and Results}

\textbf{Training Datasets.} We create multiple subsets of the original WebVid-10M dataset, of varying sizes - 10K, 20K, 40K, 80K, and 160K. The smaller datasets are a subset of the larger ones, i.e., 10K is a subset of 20K, which is a subset of 40K, and so on. This corresponds to the values of $L=\{1,2,4,8,16\}$ such that each training run contains exactly 10,000 input samples.  We conduct all training runs using 2 L40 GPUs.\\

\textbf{Evaluation.} We evaluate our finetuned models on the VideoMME video understanding benchmark. VideoMME comprises 900 videos totaling approximately 254-256 hours of content, with 2700 human-annotated multiple-choice question-answer pairs. The dataset consists of videos from six primary visual domains: knowledge, film \& television, sports competition, artistic performance, life record, and multilingual. The videos are further categorized as short ($< 2$ minutes), medium (4-15 minutes), and long (30-60 minutes). The primary metric used to evaluate video-VLMs is accuracy on the multiple choice QA task. \\

\textbf{Architectural Details}

We finetune our models for 1 epoch in each scenario. The LLM backbone is LLama-2~\cite{touvron2023llama} 7B model, vision encoder is CLIP-ViT/L-14. The video frames are naively resized to 336x336 (depends on the CLIP encoder used). All our experiments are conducted on 2 Nvidia L40s GPUs, with each training runs taking between 5 - 100 hours. We show our experiments below, with each one testing individual aspects of our method.

\subsection*{Experimental Results}


\begin{table}
  \caption{Our method compared to the baseline VideoMME under fixed compute setting (1 epoch, 10,000 training iterations).  For all methods, we start with a LLaMA-2 backbone with a frozen CLIP-ViT/L-14 visual encoder finetuned on the LLaVA-1.5 training set.}
  \vspace{2mm}
  \label{tab:ours-vs-baselines}
  \centering
  \begin{tabular}{lllll}
    \toprule
     Method     & VideoMME (short) ($\uparrow$) & VideoMME (all) ($\uparrow$) \\
    \midrule
    Image Baseline (No FT) & 34.2 & 31.7 \\
    Single-Video FT & \underline{41.7} & \underline{34.5} \\
    \midrule
    Multi-Video FT (ours)  & \textbf{44.8}  & \textbf{36.6} \\
    \bottomrule
  \end{tabular}
\end{table}








We show our main results compared to baselines in Table \ref{tab:ours-vs-baselines}. We outperform both the image-initialized VLM and standard video finetuning on the VideoMME dataset. Additionally, we show results for the setting without clustering in Table \ref{tab:no-clustering}. We note that both $L=2$ and $L=4$ outperform standard finetuning. We take the best performing values of Table \ref{tab:no-clustering} and show them with clustering in Table \ref{tab:with-clustering}.  We note that our caption enrichment strategy also performed worse than using the default captions. For $L=2$, we achieve a VideoMME score of 21.6, a 15 point drop in performance compared to using a sequence of the original WebVid10M captions.

\begin{table}
  \caption{Model performance under a fixed compute budget (1 epoch, 10,000 training iterations) with varying video-frame sampling strategies. L=16 samples 1 frame each from 16 videos per iteration (160,000 videos total). L=1 samples all 16 frames from a single video per iteration (10,000 videos total). All strategies process 160,000 total frames. Bold indicates best performance; underline indicates second best.}
  \label{tab:no-clustering}
  \centering
  \begin{tabular}{lll}
    \toprule
     Videos ($L$)    & Frames per video     & VideoMME ($\uparrow$)\\
    \midrule
     16 & 1  & 33.3\\
     8 & 2 & 34.3\\
    4 & 4 & \underline{35.4}\\
    2 & 8 & \textbf{36.6}\\
    \midrule
    1 & 16 & 34.5\\
    \bottomrule
  \end{tabular}
\end{table}

\begin{table}
  \caption{Model performance across across videos and frames using our modified K-means clustering method described in Section \ref{sec:clustering} under a fixed compute budget (1 epoch, 10,000 training iterations). We show that random selection outperforms K-means clustering for grouping input videos $V_t$.}
  \label{tab:with-clustering}
  \centering
  \begin{tabular}{llll}
    \toprule
     Videos     & Frames per video  & VideoMME (random)   & VideoMME (clustered)\\
    \midrule
    4 & 4 & 34.3 & 14.8\\
    2 & 8 & 35.4 & 31.6\\
    \midrule
    1 & 16 & 34.5 & N/A \\
    \bottomrule
  \end{tabular}
\end{table}



\subsection*{Qualitative Results}
\begin{figure*}[h]
    \centering
    \includegraphics[width=1.0\textwidth]{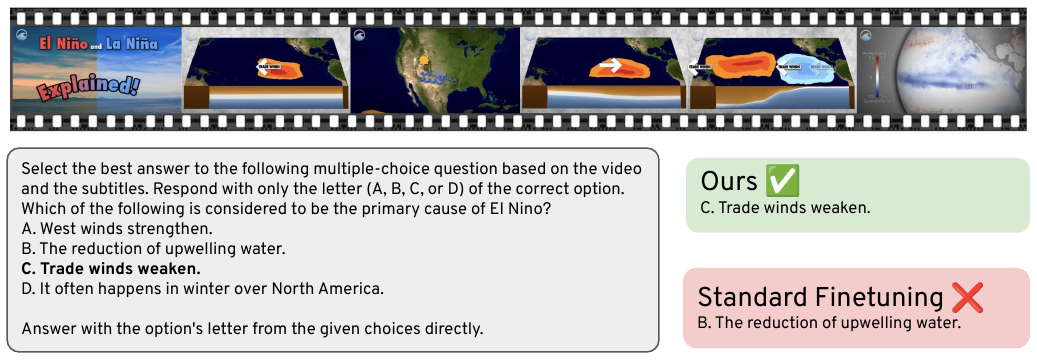}
    \caption{We show an example question from VideoMME where our model correctly identifies the correct answer. As is true across all our results, both models are evaluated using an identical set of 16 frames uniformly sampled across the video.}
    \label{fig:qual_results}
\end{figure*}

We show an example from VideoMME where our model correctly identifies the correct answer in Figure \ref{fig:qual_results}.  Often, long-video benchmarks contain videos that contain multiple disparate scenes, however our \methodname method captions these scenes well. 

\subsection*{Comparison to Single Video (Standard) Finetuning}
\begin{figure}[h]
  \centering
    \begin{subfigure}[b]{0.48\textwidth}
    \includegraphics[width=\linewidth]{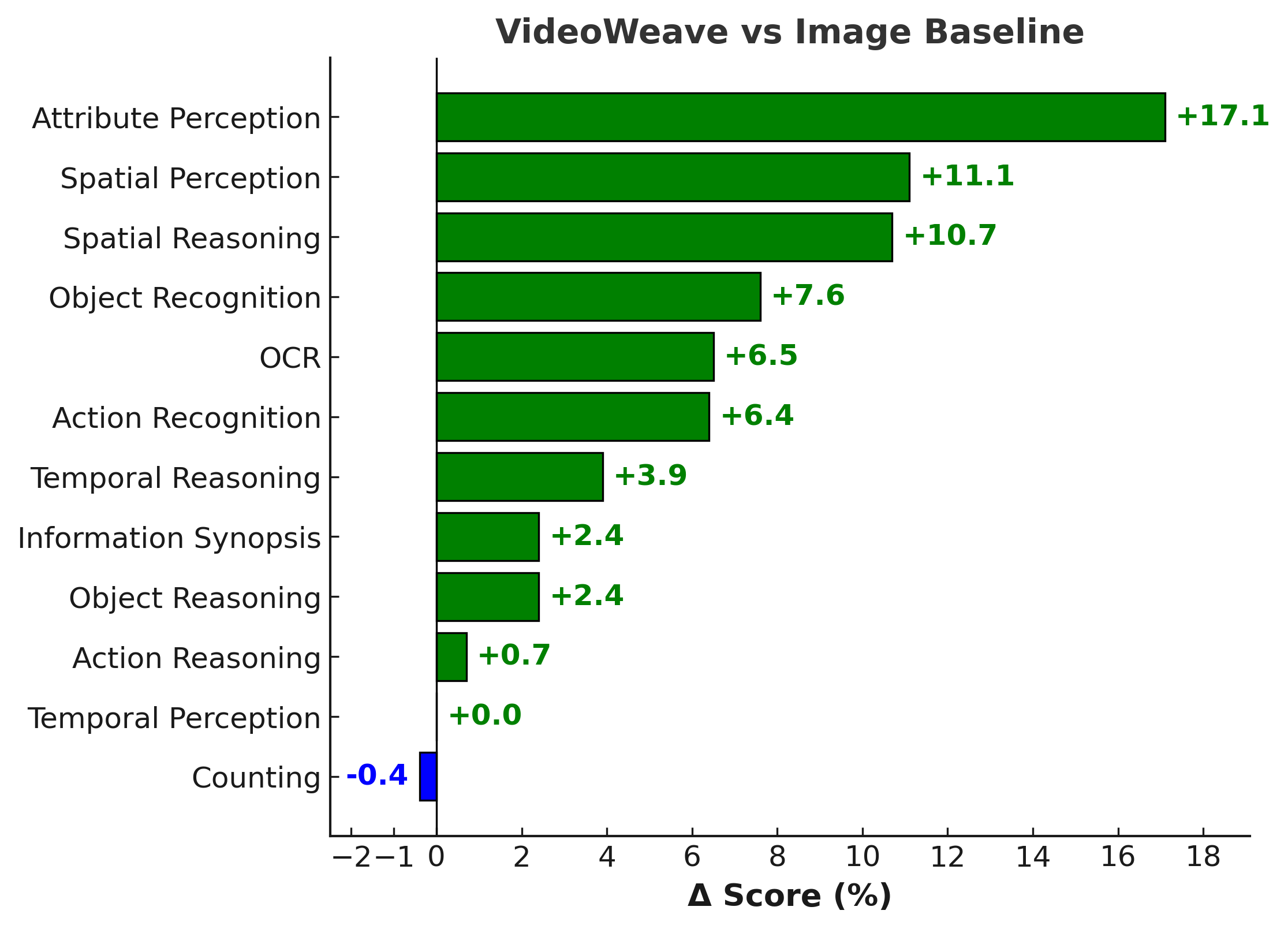}
    \caption{VideoWeave vs Image Baseline}
    \label{fig:delta_image_multi}
  \end{subfigure}
  \hfill
  \begin{subfigure}[b]{0.48\textwidth}
    \includegraphics[width=\linewidth]{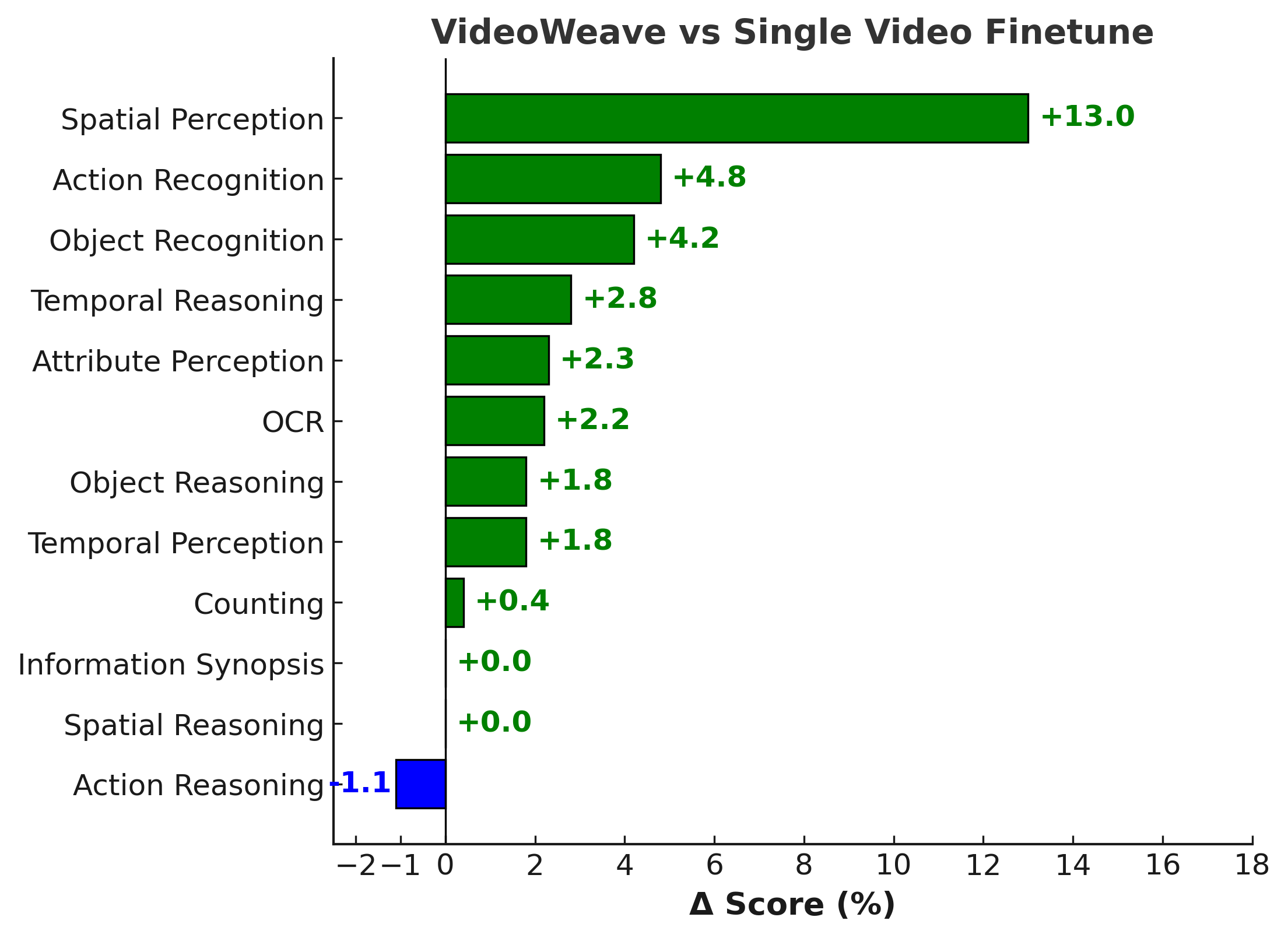}
    \caption{VideoWeave vs Single Video Finetune}
    \label{fig:delta_single_multi}
  \end{subfigure}
  \caption{Per-category performance improvements of multi-video finetuning over (a) single-video finetuning and (b) image baseline on VideoMME.}
  \label{fig:detailed-comparison}
\end{figure}

Interestingly, we find that random video splicing outperforms other approaches and is a very strong baseline for a simple strategy for improving model performance. We provide detailed per-category visualizations of our performance compared to image-level and traditional video finetuning baselines in Figure \ref{fig:detailed-comparison}.  Model performance is not significantly improved for reasoning tasks, and we hypothesize this is due to the underlying language model being shared across all models (LLaMA-2). Additionally, due to the relatively small number of frames sampled (16) without any frame selection, our model sees the greatest improvement in the short video subset of VideoMME (>3\% absolute, >7\% relative).

\section{Related Works}
\label{sec:related_works}

\subsection*{Vision-Language Modeling Architectures}

Recent advancements in vision-language models~\cite{radford2021learning, li_blip_2022, alayrac2022flamingo, yang2023dawn, zhang2023gpt4v, openai2023gpt4v} have largely adopted modular architectures that combine specialized visual encoders with large language models (LLMs). Flamingo~\cite{alayrac2022flamingo} introduced cross-modal adapters that allowed a frozen 70-billion-parameter language model to attend directly to visual tokens extracted from a ResNet-based encoder. Building upon this concept, BLIP-2~\cite{li_blip_2022} and LLaVA~\cite{liu_visual_2023} connected pretrained visual encoders, such as ViT or CLIP~\cite{radford2021learning}, to frozen LLMs through lightweight bridging modules, including a Query Transformer in BLIP-2~\cite{li_blip_2022} and a linear projection layer in LLaVA~\cite{liu2023visualinstructiontuning}. Although certain recent architectures have explored end-to-end unified models without dedicated visual encoders \cite{diaounveiling,wang2024emu3}, the modular architecture remains the predominant approach in multimodal language modeling research due to its efficiency and flexibility.

\subsection*{Video-Language Understanding and Temporal Modeling}

Expanding vision-language models from static images to videos introduces substantial challenges associated with capturing temporal information. Existing methods address this by introducing various temporal aggregation strategies prior to the language modeling stage. For instance, Valley~\cite{maaz2024video} employs a shallow transformer and pooling mechanism to aggregate temporal frame information into compact representations. Video-ChatGPT~\citep{Maaz2023VideoChatGPT} uses a similar idea but aggregates spatial tokens across frames via a LLaVA-based encoder~\citep{liu2023visualinstructiontuning}. Another notable approach, LLaVA-Video~\cite{zhang2024video}, samples frames densely at approximately one frame per second, optimizing the number of visual tokens to fit within the language model’s context window, thus retaining long-term temporal coherence. An alternative approach, exemplified by Tarsier~\cite{wang2024tarsier}, involves directly encoding frames individually with a CLIP-ViT~\citep{radford2021learningtransferablevisualmodels} encoder and subsequently modeling sequences of frame embeddings with an LLM to capture detailed temporal relationships for tasks like video captioning.

\subsection*{Benchmarks for Long Video Understanding}

Evaluating video-language models requires benchmarks designed specifically to measure long-term video comprehension and temporal reasoning. LVBench~\citep{wang2024lvbenchextremelongvideo} is one such benchmark focusing on extended video comprehension tasks across various publicly sourced domains, including television series and sports broadcasts. Similarly, Video-MME~\citep{fu2024videommefirstevercomprehensiveevaluation} provides a comprehensive framework to assess multimodal language models, covering diverse video domains to rigorously test models' temporal reasoning capabilities. HourVideo~\cite{chandrasegaran_hourvideo_2024} extends these efforts further by evaluating models on hour-long egocentric videos, encompassing challenging tasks like summarization, visual reasoning, and navigation, and highlighting performance gaps in current approaches. EgoSchema~\cite{mangalam2023egoschema} complements these efforts by providing a diagnostic benchmark built from real-world data comprising over 5,000 multiple-choice questions based on extensive egocentric video sequences, thereby emphasizing the complexity and necessity of long-form video comprehension.

\subsection*{Synthetic Video-Text Datasets}

Due to the scarcity and high annotation cost of extensive video-text datasets, recent research has turned to synthetic data generation techniques. ShareGPT4Video~\cite{chen2024sharegpt4video} is a large-scale synthetic dataset featuring video captions generated primarily by GPT-4V~\cite{openai2023gpt4v}, alongside millions more generated by the ShareCaptioner-Video model, to enhance performance on video understanding and generation tasks. Similarly, LLaVA-Video-178K~\cite{gurram2022lava} includes synthetic video instruction-following data with detailed captioning and multiple-choice question-answering tasks, enabling more effective training of video-language models. These synthetic datasets provide crucial resources that alleviate the data scarcity issue, supporting the training of more robust and capable multimodal models.

\subsection*{Efficient Training of Video-Language Models}

Despite significant progress enabled by these methods, benchmarks, and datasets, efficiently training video-language models on extensive video sequences remains challenging due to computational constraints~\cite{shin2025video, wan2023efficient, weng2024longvlm, ju2022prompting}. Existing approaches often require intensive computation, restricting scalability and practical application. Our work addresses this limitation by introducing a novel training strategy involving frame sampling from multiple videos within a single training instance, thus substantially reducing computational load. Unlike traditional single-video training strategies, our method requires no architectural modifications, providing a versatile solution for enhancing training efficiency across diverse video-language modeling tasks and scenarios.

\section{Conclusion}

In this work, we propose and study an efficient way to train VLMs for video understanding tasks called \methodname. For shorter videos with uniform content that are present in the WebVid-10M dataset, our experiments show that we can improve performance by putting together video frames from multiple different videos during training and can even use the pre-existing video captions without any further modification. \methodname achieves substantial performance gains over traditional finetuning (up to 3\% on Video-MME-Short), while not requiring significant data preprocessing or increasing modeling complexity. Thus, it can be used to facilitate training on large-scale video datasets under tight compute budgets.

That said, our work has limitations that present opportunities for future exploration. \methodname's effectiveness has been validated to help training on shorter videos with relatively uniform content, and its benefits on longer-form videos with significant temporal dynamics remain to be explored. Additionally, while our experiments on WebVid-10M and VideoMME are promising, scaling experiments across a wider range of evaluation datasets or using larger training sets would provide stronger evidence for the approach's applicability to web-scale training scenarios.  

\section{Acknowledgments}
This work was supported by a research award from Schmidt Futures, the National Science Foundation's Graduate Research Fellowship Program, and a Google Cloud Platform credit grant from Stanford Institute for Human-Centered Artificial Intelligence.

\clearpage
{
    \small
    \bibliographystyle{unsrt}
    \bibliography{references}
}

\clearpage

\end{document}